\documentclass[runningheads]{llncs}
\usepackage[T1]{fontenc}
\usepackage{graphicx,verbatim}
\usepackage{cite}
\usepackage[colorlinks=true,citecolor=blue,linkcolor=blue,urlcolor=blue]{hyperref}
\usepackage{amssymb} 
\usepackage{amsmath}
\usepackage{amsfonts}
\usepackage{booktabs}
\usepackage{multirow} 
\usepackage[most]{tcolorbox}
\usepackage{pifont}
\usepackage{adjustbox}

\begin{document}

\title{Dual-Adaptive SAM3: Hierarchical Routing over Low-Rank Expert Layers for Parameter-Efficient Medical Image Segmentation}

\titlerunning{Dual-Adaptive SAM3}

\author{
    Ying Chen\inst{1} \and
    Jinyue Li\inst{2} \and
    Kun Wang\inst{3}\textsuperscript{\textdagger} \and 
    Qiankun Li\inst{4}\textsuperscript{\textdagger} \and 
    Yang Liu\inst{3} 
}
\authorrunning{Y. Chen et al.}
\institute{
    Shenzhen Research Institute, The Chinese University of Hong Kong
    \and
    University of Science and Technology of China
    \and
    Nanyang Technological University
    \and
    Imperial Global Singapore (IGS), Imperial College London\\
    \email{kun.wang@ntu.edu.sg, q.li2@imperial.ac.uk}\\
\textsuperscript{\textdagger}Corresponding author.}
\maketitle

\begin{abstract}
The Segment Anything Model with Concepts (SAM3) heralds a new paradigm for open-vocabulary segmentation through natural language interaction, offering significant potential for medical image analysis. However, effectively adapting such a powerful vision-language model to the diverse and nuanced domain of medical imaging remains a key challenge. Naive fine-tuning is parameter-inefficient, while standard Mixture-of-Experts (MoE) methods introduce prohibitive computational overhead, limiting their clinical applicability. To address this, we propose Dual-Adaptive SAM3 (DA-SAM3), a novel framework that achieves both high segmentation accuracy and extreme parameter efficiency via a dual-adaptive specialization mechanism. Our first adaptation is task-aware: a Dynamic Expert Router (DER) that sparsely activates the most relevant experts by jointly reasoning about the visual input and the textual concept prompt, mimicking a clinical consultation process. Our second adaptation is parameter-aware: a Decomposed Parameterized Experts (DPE) design that represents each expert as a shared frozen base (inherited from the pretrained SAM3) and a lightweight trainable low-rank delta, reducing MoE parameter overhead by over \textbf{80\%}. Extensive experiments on multiple public medical segmentation benchmarks demonstrate that Dual-Adaptive SAM3 not only matches or exceeds the accuracy of fully fine-tuned SAM3 and standard MoE baselines, but also achieves a notable \textbf{5\%} gain over current state-of-the-art methods, with interpretable results validating its effectiveness. The code is available at: \url{https://github.com/Reconsider80/DA-SAM3}.
\end{abstract}

\section{Introduction}

Medical image segmentation is fundamental to diagnostics and treatment planning, enabling critical applications from tumor delineation to organ analysis. The rise of large-scale vision-language foundation models, such as the Segment Anything Model (SAM) ~\cite{kirillov2023segment} and its conceptual extension SAM3 ~\cite{carion2025sam}, introduces a paradigm shift by supporting natural language-guided segmentation (e.g., “segment the left ventricle”). This interactivity promises more intuitive clinician-AI collaboration and alleviates reliance on large annotated medical datasets ~\cite{huang2025eval}. However, deploying such general-purpose models in the highly specialized domain of medical imaging presents a critical efficiency-specialization trade-off. The heterogeneity of imaging modalities, anatomical regions, and pathological appearances demands adaptable representations ~\cite{wu2026sictta}, yet common adaptation strategies are ill-suited: full fine-tuning is parameter-inefficient and risks catastrophic forgetting of valuable pretrained knowledge, while conventional Mixture-of-Experts (MoE) models introduce prohibitive parameter and memory overhead ~\cite{wu2026accelerating}, limiting their scalability in resource-constrained clinical settings.

In pursuit of parameter efficiency, recent Parameter-Efficient Fine-Tuning (PEFT) methods like LoRA~\cite{hu2022lora} and Adapters offer a compelling alternative. However, their typical static and uniform application across the model fails to address the dynamic and hierarchical nature of medical concept understanding. Concurrently, novel efficient MoE designs like DeRS ~\cite{huang2025ders} demonstrate the potential of “upcycling” pretrained weights into conditional experts with minimal overhead. Despite these parallel advances, a crucial gap remains: how to integrate dynamic, input-conditioned routing with extreme parameter efficiency within a multimodal vision-language framework for medical segmentation. Thus, our core research question is: How can we efficiently specialize a large vision-language model like SAM3 to diverse medical concepts in a parameter-aware and task-aware manner, without sacrificing accuracy or practical deployability?

To bridge this gap, we propose Dual-Adaptive SAM3 (DA-SAM3), a novel framework that synergizes two complementary adaptive mechanisms for efficient medical image segmentation. First, a task-aware Dynamic Expert Router (DER) performs sparse, multimodal expert selection by jointly reasoning about visual content and textual concepts, emulating a context-sensitive clinical consultation. Second, a parameter-aware Decomposed Parameterized Experts (DPE) design decomposes each expert into a shared frozen based inherited from the pretrained SAM3 and lightweight low-rank “delta” matrices, inspired by efficient upcycling paradigms. Crucially, these dual mechanisms are orchestrated within a hierarchical specialization strategy, where experts at different network depths learn to specialize in coarse alignment, semantic identification, and boundary refinement, mirroring a coarse-to-fine clinical reasoning process. DA-SAM3 achieving an approximately \textbf{5\%} improvement over state-of-the-art methods but also reduces MoE parameter overhead by over \textbf{80\%}. Our contributions are threefold: 

    (1) We propose Dual-Adaptive SAM3, the first framework to integrate dynamic multimodal routing with parameter-efficient expert decomposition for adapting vision-language foundation models to medical imaging.
    
    (2) We introduce the Dynamic Expert Router (DER) for context-aware sparse expert selection, and Decomposed Parameterized Experts (DPE), a novel low-rank parameterization that drastically reduces MoE parameter overhead. 
    
    (3) Through comprehensive experiments on multiple public benchmarks, we demonstrate that our approach matches or surpasses the accuracy of standard fine-tuning and MoE baselines, while significantly improving parameter and computational efficiency providing a practical pathway toward deploying versatile, generalist AI models in clinical practice.

\section{Method}

\subsection{Overview} 

\begin{figure}[h]\centering
\includegraphics[scale=0.28]{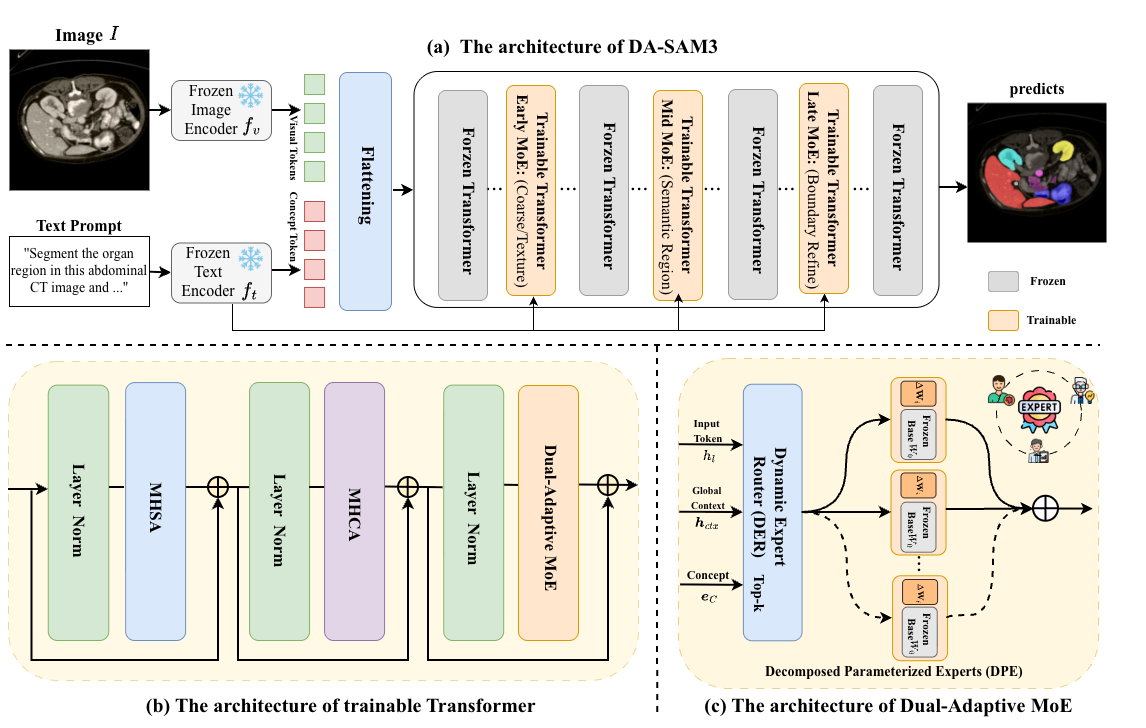}
\caption{ \textbf{(a)} The architecture of the Dual-Adaptive SAM3 framework. \textbf{(b)} The architecture of trainable Transformer. \textbf{(c)} The Dual-Adaptive MoE Layer including Dynamic Expert Router 
\textbf{(DER)} and Decomposed Parameterized Expert \textbf{(DPE)}  } .
\label{fig1}
\end{figure}

\vspace{-2em}
As shown in Fig.\ref{fig1}, we propose Dual-Adaptive SAM3 (DA-SAM3), a parameter-efficient framework for medical image segmentation. DA-SAM3 dynamically specializes the Segment Anything Model with Concepts (SAM3) via a hierarchical Mixture-of-Experts (MoE) adaptation. The core innovation is the targeted replacement of Feed-Forward Network (FFN) blocks within SAM3’s fusion module with our Dual-Adaptive MoE Layers, enabling context-aware feature specialization guided by both visual characteristics and textual prompts. DA-SAM3 adapts the concept-conditioned SAM3 architecture through three main components: (1) a frozen image encoder $f_v$ extracting visual features $V$; (2) a frozen text encoder $f_t$ providing concept embeddings $e_c$; and (3) a trainable Transformer-based fusion module. To maintain parameter efficiency and preserve pre-trained general knowledge, $f_v$ and $f_t$ remain entirely frozen. We exclusively fine-tune the newly inserted Dual-Adaptive MoE layers and Layer Normalization (LN) parameters within the fusion module. All other components, including the multi-head self-attention (MHSA) layers and Multi-Scale Cross-Attention (MSCA) layers, are kept frozen. This selective optimization strategy ensures dynamic domain adaptation while minimizing the trainable parameter footprint.

\subsection{Hierarchical Fusion Decoder}
The Hierarchical Fusion Decoder is a lightweight Transformer-based module designed to ground the linguistic [SEG] query into the visual domain. Unlike standard decoders that treat all layers uniformly, DA-SAM3 strategically substitute standard FFNs with \textbf{Dual-Adaptive MoE (DA-MoE)} layers at critical depths $\{L/6, L/4, L/2\}$. This hierarchical placement is not merely a structural choice but is intentionally designed to emulate the \textbf{multi-stage clinical reasoning process} typically employed by radiologists:

\textbf{Early Stage (Coarse Global Alignment, $L/6$ )}: 
Mirroring the "initial survey" in clinical reading, the MoE leverages global context $\mathbf{h}_{c t x}$ to align low-level visual primitives with the input concept.

\textbf{Mid Stage (Semantic Structure Identification, $L/4$ }): 
At the intermediate depth, the model transitions to "pattern recognition." The experts focus on integrating multi-modal features to resolve complex morphological variations and identify organ-specific textures. This stage ensures the model correctly distinguishes between adjacent structures with similar intensities. 

\textbf{Late Stage (Precision Boundary Refinement, $L / 2$ )}: Equivalent to the "fine-grained contouring" required for surgical planning, these experts operate on high-resolution feature maps. The routing mechanism prioritizes experts capable of sharpening boundaries and resolving regional ambiguities, ensuring the final mask captures subtle pathological extensions or thin tissue interfaces.

The module processes the visual feature map $V \in \mathbb{R}^{H \times W \times D_v}$ and concept embedding $e_c \in \mathbb{R}^{D_t}$ into a unified sequence $\mathbf{X}_0$:

\begin{equation}
\mathbf{X}_0 = [\mathbf{V}{\text{tok}} ; \mathbf{C}{\text{tok}} ; \mathbf{S}{\text{tok}}] + \mathbf{P} \in \mathbb{R}^{(N_v + 2) \times D},\end{equation}
where $\mathbf{V}_{\text {tok }}, \mathbf{C}_{\text {tok }}$, and $\mathbf{S}_{\text {tok }}$ represent flattened visual features, projected concept embeddings, and a learnable segmentation query, respectively. $\mathbf{P}$ denotes learnable positional encodings. The sequence $\mathbf{X}_0$ is propagated through $L$ layers. 
\vspace{-2.0em}

\subsection{Dual-Adaptive MoE Layer}

\vspace{-0.5em}
We replace the standard FFN in selected layers of decoder with our proposed Dual-Adaptive MoE Layer. For a target layer $l$ receiving input features $\mathbf{H}_l$, this layer operates as follows:
% \vspace{-1.0em}
\textbf{Dynamic Expert Router (DER)}
The router generates sparse, token-wise gating weights by conditioning on a dual signal: a global domain-context vector and the local token features.
Domain-Context Vector $\mathbf{h}_{ctx}$ : To inform the router about the global image domain and its relevance to the concept, we compute a compact summary. The concept token $\mathbf{C}_{tok}$ attends to a spatially pooled version of the visual features:

\begin{equation}
\mathbf{h}_{ctx}=\operatorname{CrossAttn}\left(\mathbf{C}_{t o k}, \operatorname{Pool}(\mathbf{V})\right).
\end{equation}

Routing Score Calculation: For the $j$-th token $\mathbf{h}_l^j$ in $\mathbf{H}_l$, the score for expert $i$ is computed by fusing the token's content, the global context, and the original concept semantics:

\begin{equation}
s_i^j=\mathbf{W}_r \cdot\left[\mathbf{h}_l^j ; \mathbf{h}_{c t x} ; \operatorname{StopGrad}\left(e_c\right)\right],
\end{equation}
where $\mathbf{W}_r$ is a learnable projection and StopGrad ensures the router does not distort the frozen concept embedding.
Sparse Top-k Gating: A sparse gating function selects the top-$k$ experts per token for efficient computation, producing routing weights $p_i^j$. A sparse gating function selects the top- $k$ experts (we use $k=2$ ) per token for efficient computation, producing routing weights $p_i^j$.
\textbf{Decomposed Parameterized Expert (DPE)}
Inspired by DeRS, we parameterize each expert $E_i$ not as a full dense network, but as a shared base weight plus a lightweight expert-specific delta. The original pre-trained FFN weights $\mathbf{W}_0$ from the corresponding SAM3 layer are retained as a shared, frozen base.
Each expert $i$ is defined by a lightweight, learnable low-rank delta:

\begin{equation}
\Delta \mathbf{W}_i=\mathbf{A}_i \mathbf{B}_i^{\top},\quad \mathbf{A}_i \in \mathbb{R}^{D \times r},  \mathbf{B}_i \in \mathbb{R}^{D_{f f} \times r}, r \ll D.
\end{equation}

The expert's operation is: 

\begin{equation}
E_i(\mathbf{x})=\left(\mathbf{W}_0+\Delta \mathbf{W}_i\right) \mathbf{x}.
\end{equation}

The output for token $j$ is: 

\begin{equation}
\mathbf{y}_l^j=\sum_{i \in \mathrm{TopK}} p_i^j \cdot \operatorname{GELU}\left(E_i\left(\mathbf{h}_l^j\right)\right).
\end{equation}

This design introduces minimal new parameters while enabling dynamic specialization.

\subsection{Two-Stage Specialization}

We employ a two-stage training strategy to ensure stable expert specialization and effective routing.
\textbf{Warm-up Stage (Expert Specialization)}: We initialize the low-rank delta matrices $\left\{\mathbf{A}_i, \mathbf{B}_i\right\}$ to zero and the router randomly. The model is first trained on a diverse medical segmentation dataset with a standard segmentation loss $\mathcal{L}_{\text {seg }}$. An auxiliary Load Balancing Loss $\mathcal{L}_{\text {balance }}$ is added to prevent router collapse and ensure equitable expert utilization across the batch.
\textbf{Fine-tuning Stage (Routing Calibration)}: We then fine-tune the router parameters (with experts fixed) on a broader dataset that includes challenging or composite concepts. This stage enhances the model's open-vocabulary generalization.
The final training objective is:
\begin{equation}
\mathcal{L}_{\text {seg }}=\mathcal{L}_{\text {Dice}}+\mathcal{L}_{\text {Focal Loss}},
\mathcal{L}=\mathcal{L}_{\text {seg }}+\lambda_1 \mathcal{L}_{\text {balance }}+\lambda_2 \mathcal{L}_{\text {sparse},}
\end{equation}
where $\mathcal{L}_{\text {sparse }}$ is a regularization term encouraging the sparsity of the routing weights.

\section{Experiments}

\subsection{Datasets} We evaluate DA-SAM3 on four widely adopted public datasets spanning cardiac and abdominal imaging. The \textbf{Synapse} dataset ~\cite{fang2020multi} comprises 30 abdominal CT scans, with 18 allocated to training and 12 to testing, featuring eight abdominal organ labels. The \textbf{MMWHS} dataset ~\cite{zhuang2016multi} contains 20 cardiac CT volumes, which we split into 16 training and 4 testing instances following the original protocol. The \textbf{BTCV} dataset ~\cite{landman2015miccai} provides 30 annotated CT volumes covering 13 abdominal structures; we adopt its standard split of 24 training and 6 testing cases. The \textbf{ACDC} dataset ~\cite{bernard2018deep} contributes 150 cardiac MRI acquisitions, all utilized with their predefined training-test partitions. For all experiments, we adhere to the official data splits to ensure fair comparison with prior work. 

\subsection{Implementation Details and Evaluation Metrics} 

To facilitate training, we apply various data augmentation techniques, including flipping, rotation, scaling, and intensity shifting. All images, except those from the Synapse CT dataset, are resized to 256$\times$256, while Synapse CT images are resized to 224$\times$224. The model is trained with a batch size of 8 using the AdamW optimizer, and weight decay = 0.1. The hyper-parameters $\lambda_1$ and $\lambda_2$ were empirically set to $0.01$ and $0.001$, respectively. This configuration follows the established practice in sparse MoE training, ensuring a balanced expert utilization and sharp routing decisions without compromising the convergence of the primary segmentation objective $\mathcal{L}_{seg}$. The learning rate is set to 0.0005, and a warmup strategy is employed to ensure stable convergence during the early stages. In the MoE configuration, we set the number of experts to 4 and the top-k value to half of the total feature count.

\subsection{Comparison with State-of-the-Art Methods}

To validate the efficacy of the proposed DA-SAM3, we conducted extensive benchmarks against contemporary SOTA SAM-based variants and task-specific architectures across four standardized datasets. These SAM-derived frameworks are bifurcated into prompt-guided and prompt-agnostic categories. For the former, a single point, stochastically sampled from the gold-standard mask, serves as the input prompt.Quantitative evidence in Table \ref{tab:Com1} reveals that DA-SAM3 establishes a new performance ceiling, markedly eclipsing existing SAM adaptations in both Dice Similarity Coefficient (DSC) and Hausdorff Distance (HD). Furthermore, as detailed in the upper section of Table \ref{tab:Com1}, our model consistently outperforms domain-specialized networks, underscoring its robust competitive edge in medical scenarios.

\begin{table*}[h]
\centering
\caption{Comparison with state-of-the-art methods on Synapse CT, MMWHS, BTCV, and ACDC.}
\scalebox{0.67}{
\begin{tabular}{@{}c|c|cc|cc|cc|cc@{}}
\toprule
\multirow{2}{*}{Category} & \multirow{2}{*}{Method} & 
\multicolumn{2}{c|}{Synapse CT} & \multicolumn{2}{c|}{MMWHS} & 
\multicolumn{2}{c|}{BTCV} & \multicolumn{2}{c}{ACDC} \\
\cmidrule(lr){3-4} \cmidrule(lr){5-6} \cmidrule(lr){7-8} \cmidrule(l){9-10}
& & DSC $\uparrow$ & HD $\downarrow$ & DSC $\uparrow$ & HD $\downarrow$ & 
DSC $\uparrow$ & HD $\downarrow$ & DSC $\uparrow$ & HD $\downarrow$ \\
\midrule

\multirow{5}{*}{Task-specific} 
& nnU-Net ~\cite{isensee2021nnu} & 79.89 & 28.520 & 87.55 & 17.720 & 72.67 & 15.720 & 91.54 & 1.086 \\
& TransUNet ~\cite{chen2021transunet} & 79.95 & 11.580 & 88.47 & 24.310 & 77.67 & 9.498 & 88.10 & 1.538 \\
& Swin-UNETR ~\cite{hatamizadeh2021swin}  & 80.58 & 15.460 & 88.92 & 14.310 & 78.11 & 7.405 & 89.74 & 1.239 \\
& MedNeXt ~\cite{roy2023mednext}  & 82.69 & 11.980 & 88.55 & 14.950 & 80.81 & 7.379 & 90.88 & 1.129 \\
& Swin-UMamba ~\cite{liu2024swin}  & 83.48 & 8.140 & 88.91 & 15.060 & 80.59 & 5.910 & 90.39 & 1.253 \\
\midrule

\multirow{4}{*}{Prompt-free SAM}
& SAMed ~\cite{zhang2023customized}  & 80.42 & 10.770 & 87.050 & 25.320 & 71.23 & 9.010 & 88.83 & 1.429 \\
& AutoSAM ~\cite{shaharabany2023autosam} & 81.61 & 10.170 & 88.71 & 12.990 & 75.77 & 7.693 & 72.05 & 3.248 \\
& H-SAM ~\cite{cheng2024unleashing} & 80.27 & 13.170 & 87.33 & 14.110 & 72.81 & 7.037 & 88.38 & 1.410 \\
& MoE-SAM ~\cite{li2025moe} & 84.71 & 8.756 & 89.38 & 13.670 & 76.82 & 5.637 & 91.89 & 1.064 \\
\midrule

\multirow{7}{*}{Prompt-based SAM} 
& SAM ~\cite{kirillov2023segment}  & 64.94 & 39.830 & 82.11 & 46.940 & 63.84 & 20.360 & 75.15 & 4.311 \\
& MedSAM ~\cite{ma2024segment}  & 72.45 & 20.430 & 84.53 & 55.940 & 69.14 & 18.490 & 82.11 & 3.720 \\
& MSA ~\cite{wu2025medical}  & 77.13 & 25.340 & 85.50 & 35.680 & 72.31 & 17.51 & 83.01 & 2.715 \\
& SAMUS ~\cite{lin2024beyond}  & 70.55 & 43.650 & 83.98 & 30.740 & 65.12 & 24.58 & 69.66 & 5.559 \\
& DeSAM ~\cite{gao2024desam}  & 76.77 & 9.704 & 81.54 & 16.340 & 68.08 & 7.263 & 67.26 & 5.391 \\
& SAM-Med2D ~\cite{ye2023sa}  & 66.96 & 22.850 & 81.42 & 59.660 & 53.64 & 23.630 & 80.02 & 4.587 \\
& SAM3~\cite{carion2025sam}& 80.75 & 15.120 & 85.10 & 18.100 & 72.24 & 8.080& 85.03& 5.280 \\
\hline

& \textbf{DA-SAM3 (Ours)} & \textbf{85.12} & \textbf{8.425} & \textbf{90.06} & \textbf{12.880} & \textbf{77.35} & \textbf{5.587} & \textbf{91.93} & \textbf{1.064} \\
\bottomrule
\end{tabular}}
\label{tab:Com1}
\end{table*}

\begin{figure}[h]\centering
\includegraphics[scale=0.18]{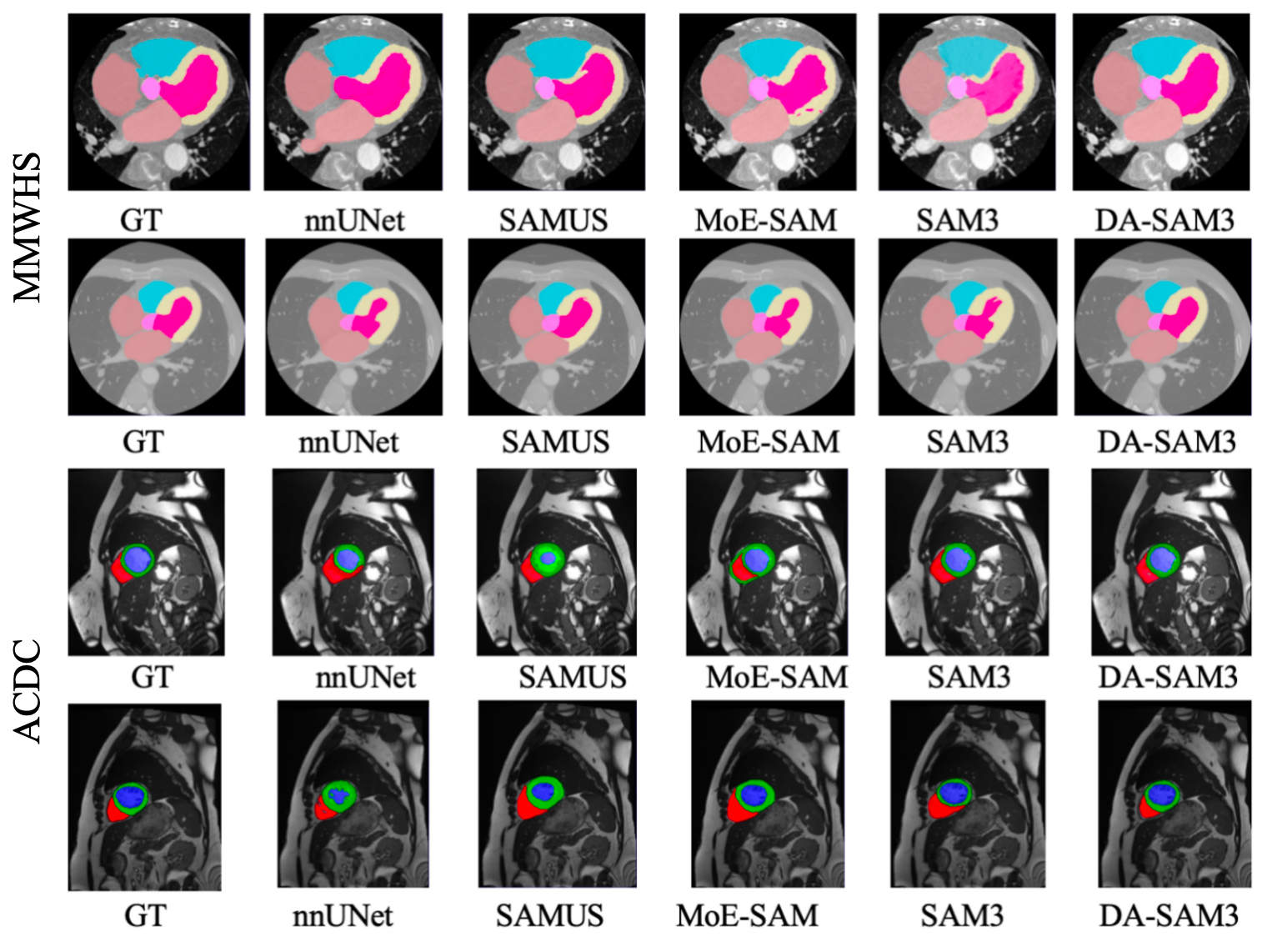}
\caption{Visual comparison of segmentation results.} 
\label{fig2}
\end{figure}

As depicted in Fig. \ref{fig2}, our method exhibits precise alignment with actual organ structures, thereby validating the efficacy of our location-aware design. In contrast, both MoE-SAM and SAM3 frequently misclassify different organs, underscoring the superiority of our Dual-Adaptive MoE Layer strategy.

\subsection{Ablation Study}

\begin{table*}[h]
\centering
\caption{Ablation study of module's key component on three datasets.}
\scalebox{0.7}{
\begin{tabular}{c|c|c|c|c|c|c|c}
\hline
\multirow{2}{*}{Category} & \multirow{2}{*}{Method} & \multicolumn{2}{|c|}{Synapse CT} & \multicolumn{2}{|c}{MMWHS} & \multicolumn{2}{|c}{ACDC} \\
\cline{3-8}
 &  & DSC $\uparrow$ & HD $\downarrow$ & DSC $\uparrow$ & HD $\downarrow$ & DSC $\uparrow$ & HD $\downarrow$ \\
\hline
\multirow{1}{*}{Baseline} & SAM3 & 80.75 & 15.120 & 85.10 & 18.100 & 72.24 & 8.080 \\
\hline
\multirow{1}{*}{Fusion Strategy Variants}
 & SAM3+Concat & 80.82 & 15.090 & 88.18 & 18.130 & 73.23 & 8.106 \\
\hline
\multirow{4}{*}{Fine-tuning Strategy Variants}
& SAM3+LoRA & 82.82 & 14.342 &83.28  &15.482  & 73.35 &  7.231\\
& SAM3+Standard MoE & 82.28 & 14.090 & 85.18 & 16.630 & 73.03 & 7.480 \\
 & Ours w/o DER (Random Route) & 82.75 & 12.120 & 86.10 & 14.100 & 73.24 & 7.080 \\
 & Ours w/o DPE (Full Expert) & 83.55 & 11.700 & 86.60 & 13.700 & 74.55 & 6.102 \\
\hline
\multirow{1}{*}{Ours} 
 & DA-SAM3 (Ours Full) & $\mathbf{85.12}$ & $\mathbf{8.425}$ & $\mathbf{90.06}$ & $\mathbf{12.880}$ & $\mathbf{77.35}$ & $\mathbf{5.587}$ \\
\hline
\end{tabular}}
\label{tab:Abl}
\end{table*}

Table~\ref{tab:Abl} presents ablation results on three datasets. Compared to the standard SAM3 baseline, our full model achieves substantial gains, notably improving DSC by 4.37\% on Synapse CT. While SAM3+LoRA and Standard MoE offer incremental improvements, they fall short of DA-SAM3’s performance. Specifically, our method reduces the HD score from 15.120mm to 8.425mm on Synapse, demonstrating that targeted expert specialization at the semantic bottleneck is more effective than global low-rank adaptations or unconstrained expert layers for capturing precise anatomical boundaries. The SAM3+Concat (where visual and text tokens are simply concatenated without hierarchical interaction) variant shows limited efficacy, confirming that simple feature aggregation cannot resolve complex cross-modal misalignments. Furthermore, replacing our DER with Random Routing leads to a significant performance drop. This underscores the router's critical role in utilizing clinical concept embeddings to dynamically activate task-relevant experts. Ablating the DPE results in inferior performance across all metrics. This validates that our parametric decomposition acts as a robust regularizer, preventing expert over-parameterization while maximizing the model's capacity to resolve ambiguous clinical textures.

\section{Conclusion}

In this paper, we introduce Dual-Adaptive SAM3 (DA-SAM3), a novel framework that achieves parameter-efficient medical image segmentation through hierarchical routing over low-rank expert layers. DA-SAM3 fundamentally rethinks adaptation by expanding a single, static adapter into multiple, dynamically selectable expert adapters, enabling the model to perform input-conditioned, coarse-to-fine reasoning that mimics clinical decision-making. A Dynamic Expert Router (DER) that sparsely activates the most relevant experts by jointly reasoning about visual content and textual concepts, and Decomposed Parameterized Experts (DPE) that decompose each expert into a shared frozen base and lightweight low-rank deltas. Extensive experiments on four public benchmark datasets demonstrate that DA-SAM3 not only matches but often surpasses the accuracy of fully fine-tuned models and standard MoE baselines, while maintaining extreme parameter efficiency.

\begin{credits}
\subsubsection{\ackname} This research is part of the IN-CYPHER programme and is supported by the National Research Foundation, Prime Minister's Office, Singapore under its Campus for Research Excellence and Technological Enterprise (CREATE) programme.

\subsubsection{\discintname}
The authors have no competing interests in the paper.
\end{credits}

\bibliographystyle{plain}
\bibliography{ref}
\end{document}